\documentclass[11pt]{article}

\usepackage[a4paper,margin=2.5cm]{geometry}
\usepackage[T1]{fontenc}
\usepackage[utf8]{inputenc}
\usepackage{lmodern}
\usepackage{microtype}
\usepackage{authblk}

\usepackage[numbers,sort&compress]{natbib}

\usepackage{graphicx}%
\usepackage{multirow}%
\usepackage{amsmath,amssymb,amsfonts}%
\usepackage{amsthm}%
\usepackage{mathrsfs}%
\usepackage[title]{appendix}%
\usepackage{xcolor}%
\usepackage{textcomp}%
\usepackage{booktabs}%
\usepackage{float}
\usepackage[hidelinks]{hyperref}

\graphicspath{{../}{./}}

\newcommand{\bmhead}[1]{\subsection*{#1}}
\newcommand{\backmatter}{}
\let\botrule\bottomrule

\theoremstyle{plain}%
\theoremstyle{definition}%

\title{Skillful Data-Driven Subseasonal Soil Moisture Forecasting: Prospects and Limits for Flash Drought Prediction}

\author[1]{Noelia Otero\thanks{Corresponding author: \href{mailto:noelia.otero.felipe@hhi.fraunhofer.de}{noelia.otero.felipe@hhi.fraunhofer.de}}}
\author[1]{Atahan Özer}
\author[2]{Miguel-Ángel Fernández-Torres}
\author[1]{Jackie Ma}

\affil[1]{Applied Machine Learning Group, Fraunhofer Institute for Telecommunications, Heinrich Hertz Institute, Berlin, Germany}
\affil[2]{Department of Signal Theory and Communications, Universidad Carlos III de Madrid (UC3M), Madrid, Spain}

\date{}

\begin{document}

\maketitle

\begin{abstract}
Despite substantial progress in short-to-medium-range weather forecasting, predicting high-impact events such as flash droughts remains a key challenge for both early warning operations and physically-based subseasonal-to-seasonal (S2S) prediction systems. Here we demonstrate that, for S2S soil-moisture forecasting over Europe, forecast skill depends as much on how the prediction problem is formulated as on the forecasting model itself. Using a Vision Transformer-based architecture with dual-pathway temporal and spatial attention, we show that residual learning is essential to outperform persistence. This advantage is realized only when forecasting root-zone soil moisture in physical units rather than standardized anomalies, revealing that the target representation itself constrains predictability. A probabilistic extension via quantile-head fine-tuning further provides well-calibrated predictive distributions. Benchmarked against deep-learning and operational ECMWF S2S baselines over 2021–2022, our model achieves the highest deterministic and probabilistic skill at all lead times and reliably detects anomalously dry root-zone states (below the 20th percentile). Yet flash drought \emph{onset}, defined by multi-pentad intensification criteria, remains a fundamental challenge shared across all current S2S systems. These findings advance data-driven S2S soil-moisture forecasting while highlighting the remaining challenge of predicting rapid drought development.
\end{abstract}

\noindent\textbf{Keywords:} flash drought, soil moisture, subseasonal-to-seasonal forecasting, deep learning, Vision Transformer, early warning

\section{Introduction}\label{sec:intro}

Traditional droughts are understood as slow-onset phenomena that may persist for months or even years. Flash droughts (FDs) represent a distinct class of drought characterized by unusually rapid intensification \citep{Osman2022,Otkin2018}. This rapid onset is associated with abrupt increases in drought severity and can lead to substantial impacts, particularly in agriculture and natural ecosystems \citep{Osman2022, Otkin2022, Tyagi2022}. Due to complex land--atmosphere interactions and their fast development, FDs remain difficult to detect and predict \citep{Tyagi2022}.

To characterize these dynamics, percentile-based criteria \citep{Yuan2019} and indices such as the Flash Drought Intensity Index (FDII; \citep{Otkin2018}) have been proposed. These indices aim to represent rapid intensification at weekly to pentad timescales, typically by combining drought severity thresholds with rates of change \citep{Tyagi2022, Otkin2018}. However, despite these advances, there is no universally accepted definition of flash drought, and different formulations can lead to substantially different event characterizations and evaluation outcomes.

This lack of consistency, combined with the limited skill of current numerical weather prediction (NWP) systems at subseasonal-to-seasonal (S2S) timescales, motivates the need for improved data-driven approaches \cite{He2025}. In particular, traditional NWP systems often struggle to provide sufficiently early and reliable signals of rapid land-surface drying events \citep{Otkin2022}.

Recent efforts have therefore increasingly explored machine learning methods for post-processing NWP outputs and directly predicting drought-related variables \citep{Xu2024, Lesinger2025, Otero2025}. Concurrently, data-driven models have shown promising performance for FD prediction and identification \citep[e.g.][]{Zhang2022, Jacob2025, Foroumandi2024, Lorenz2024}, with both classical machine learning (ML) and deep learning (DL) approaches demonstrating the ability to capture nonlinear land--atmosphere interactions and rapid intensification patterns. In particular, a two-part study by \citet{Edris2025partI,Edris2025partII} demonstrated that while classical ML methods already achieve robust performance in identifying FD characteristics, neural network-based approaches can further enhance skill by better representing nonlinear relationships and the rapid intensification phase, albeit with reduced interpretability and greater sensitivity to training data and varying FD definitions.

DL approaches have increasingly focused on soil moisture as a key predictive target due to its central role in land--atmosphere coupling and its direct relevance to drought evolution \cite{Seneviratne2010}. For instance, \citet{Zhao2025} introduced an interpretable Transformer-based framework that improves FD prediction skill while explicitly capturing nonlinear interactions and providing insights into key drivers such as temperature, radiation, and soil moisture memory. Similarly, \citet{Yang2026} showed that ensembles of spatiotemporal DL models can forecast S2S soil moisture and identify FD drivers across diverse climate regimes through post-hoc explainability methods. These studies highlight the importance of soil moisture as both a predictive target and a physically consistent variable for improving FD monitoring and forecasting. A recent study by \citet{Edris2026} explored the capabilities of the DroughtFormer, a variant of the CrossFormer \citep{wang2022crossformer} within the CREDIT framework\citep{schreck2025credt}, to predict a range of drought-related variables, including soil moisture over Africa. Their model delivered stable forecasts out to 90-day lead times, outperforming or matching climatology for most variables, though it struggled to predict precipitation and the magnitude of anomalies. Notably, their evaluation relies primarily on a climatological baseline, without comparison against operational S2S forecast systems, making it difficult to fully benchmark the reported skill against current state-of-the-art forecasting systems.

We introduce SMCast, a dual-pathway Vision Transformer (ViT) \citep{Dosovitskiy2021}
with residual learning for skillful S2S soil moisture forecasting over Europe. A ViT
is well suited to this problem because soil moisture variability reflects both
large-scale atmospheric forcing and persistent land-surface memory \citep{Dunkl2026}. Atmospheric
circulation patterns can span thousands of kilometres, creating non-local spatial dependencies that self-attention can represent directly through data-dependent interactions between spatial tokens. At the same time,
root-zone soil moisture has memory timescales ranging from weeks to several months,
with the relevant timescale depending on the land-surface state
\citep{Ghannam2016, Wu2004, Seneviratne2010}. Temporal self-attention allows each input
day to contribute directly to the forecast, rather than requiring information from
distant time steps to pass through a sequence of recurrent updates. The balance between
local soil-moisture memory and atmospheric forcing also varies across regions and with season \citep{Seneviratne2010}. This motivates our dual-pathway design, which processes temporal and spatial dependencies separately and combines them through a learnable gate, allowing their relative
contributions to vary with context. Finally, full joint space--time attention would
scale quadratically with $T_{\text{in}} \times H \times W$ and is computationally
prohibitive for 90 daily frames over the European domain. The divided space--time
formulation \citep{Bertasius2021} retains both temporal and spatial attention while
substantially reducing this computational cost.

\begin{figure}[H]
\centering
\includegraphics[width=\linewidth]{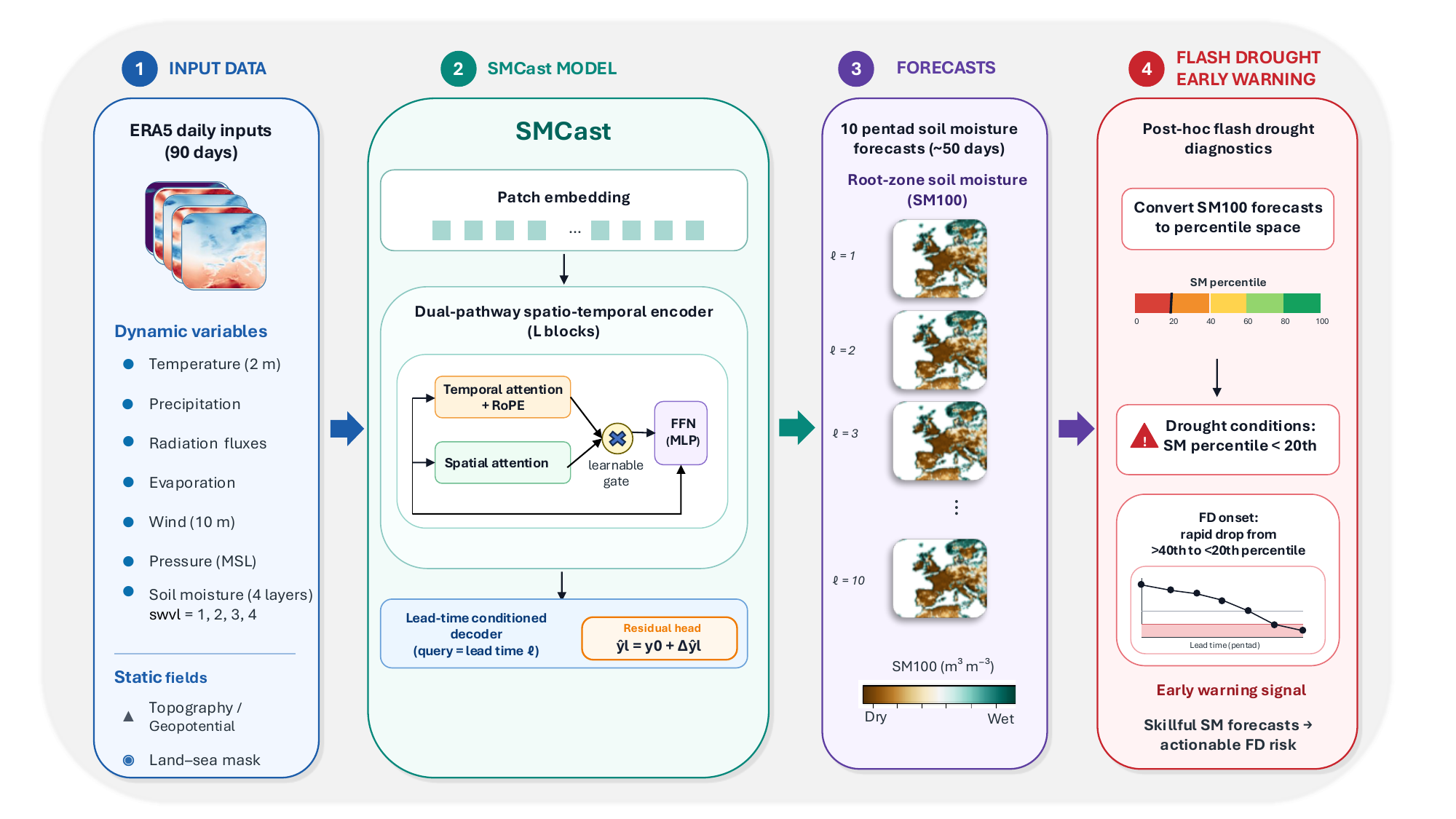}
\caption{Schematic overview of the proposed SMCast framework. (1) The end-to-end processing pipeline maps a historical sequence of $T_{\text{in}} = 90$ daily ERA5 atmospheric and land-surface fields to future subseasonal soil moisture trajectories. (2) Input frames are converted into structured sequences via non-overlapping patch embedding with spatial positional encodings. The spatiotemporal encoder features a stack of $L$ factorized (divided) space–time attention blocks, where relative temporal distances are dynamically handled using Rotary Position Embeddings (RoPE). Within each block, parallel temporal and spatial self-attention pathways isolate local land-surface memory and regional hydroclimatic co-variability, respectively, before being integrated via a learnable fusion gate. (3) The lead-time conditioned decoder utilizes a shared projection head combined with horizon-specific learnable embeddings to directly generate multi-horizon target forecasts for all $T_{\text{out}} = 10$ future pentad lead times simultaneously. Absolute target values are reconstructed using a residual learning formulation ($\hat{y}_\ell = \Delta\hat{y}_\ell + y_0$) relative to the most recent observed soil moisture state ($y_0$). (4) Flash drought events and standardized anomalies are subsequently diagnosed post-hoc by transforming the forecasted SM100 fields into percentile space.}\label{fig:framework}
\end{figure}

Beyond the model itself, we prioritize problem definition over architectural design to isolate the effects of prediction formulation and target representation. Figure~\ref{fig:framework} provides a schematic overview of the full pipeline and a detailed description can be found in Section ~\ref{subsec:model}. We show that the prediction formulation is critical: predicting deviations from the last observed state (i.e., residual learning) rather than the absolute target is required to surpass the persistence barrier. This effect is especially pronounced when the target is root-zone soil moisture in its original scale (herein referred to as SM100; i.e., representing the 0-100 cm layer), compared to standardized anomaly (SMA) representations. We further extend the deterministic framework with a probabilistic quantile-head (QH) decoder that provides well-calibrated uncertainty estimates without retraining the encoder. Furthermore, we introduce a post-hoc FD early warning pathway that maps SM100 forecasts into percentile space, showing that skillful soil moisture prediction can translate into meaningful FD signals at S2S timescales. The evaluation period (2021--2022) includes the 2022 European drought \citep{JRC2022}, providing a stringent test of model performance under extreme conditions.

\section{Results}\label{sec:result}

We evaluate SMCast in terms of (i) deterministic forecast skill for soil moisture, (ii) probabilistic calibration of the quantile forecasts, and (iii) its ability to support FD early warning through post-hoc diagnostics. The deterministic assessment is organized into two stages, motivated by the available evaluation samples. We first benchmark SMCast against three DL baselines with different inductive biases, evaluated on the full 2021--2022 test set. To ensure the comparison is robust to initialization, each DL model was trained with five random seeds, and results are reported as the mean and across-seed standard deviation. Note that this multi-seed protocol is applied only to the DL baseline comparison. The ablation experiments (input window, SM100 vs. SMA, direct vs. residual forecasting) and the comparison with operational forecasts use the seed-42 SMCast checkpoint, given that across-seed performance variations are negligible ($\pm 0.0008$ mean MAE).

We then compare SMCast, which showed the strongest skill among the DL baselines, against persistence, climatology, and the ECMWF S2S forecasts, including the quantile-mapping bias-corrected variant, ECMWF S2S~(BC). Given that ECMWF S2S forecasts are issued only biweekly (Mondays and Thursdays), ERA5 and all model-based predictions are subselected to the corresponding S2S initialization dates for this second stage, ensuring a consistent comparison on a common set of valid times. Since the ECMWF S2S forecasts extend only to day 46, this second stage is restricted to the first nine pentad lead times (days 1--45) to ensure a complete pentad dataset across all evaluated systems.

\paragraph{Comparison with deep learning baselines}\label{subsec:dlbench}

As illustrated in Table~\ref{tab:perlead_mae} (the corresponding per-lead RMSE can be found in Supplementary Table~S6), SMCast outperforms the other models at every lead time, confirming that, at matched capacity and with identical inputs (see Supplementary Text~3), the factorized dual-pathway temporal/spatial attention outperforms convolutional, recurrent, and cross-scale attention designs. Across five random seeds, the CrossFormer is the strongest baseline in average mean absolute error (MAE) ($+11.4\%$ over persistence), marginally ahead of the fully convolutional U-Net ($+10.4\%$), with U-Net leading only at the first two lead times; both clearly outperform persistence, underscoring the value of learned spatio-temporal predictors for S2S soil moisture. The recurrent E3D-LSTM, in contrast, fails to surpass persistence ($-20.8\%$) and is both the weakest and the least stable data-driven model (mean MAE $0.2402 \pm 0.0266$ across seeds, versus $\le 0.007$ for the other models) despite its larger parameter count (see Supplementary Text~3). SMCast's performance advantage is robust to initialization (overall MAE $0.1696 \pm 0.0008$) and statistically significant against every baseline (paired $t$-test on per-seed mean MAE: $p=0.001$ vs. CrossFormer, $p=0.033$ vs. U-Net, $p=0.004$ vs. E3D-LSTM). This indicates that purely recurrent video-prediction architectures are poorly suited to this slowly varying, persistence-dominated target. Interestingly, these results indicate that predictability is largely driven by the strong persistence and spatial coherence of root-zone soil moisture \citep{He2025}, enabling architectures such as the CNN-based U-Net and the cross-scale CrossFormer to achieve strong forecasting skill primarily through largely spatial representations, despite their relatively simple temporal modeling. At matched capacity and with identical inputs, this comparison establishes that the factorized dual-pathway design outperforms convolutional, recurrent, and cross-scale alternatives. However, it cannot by itself attribute this improvement to any specific component of SMCast, as the architectures differ in several respects. Thus, we quantify the individual contributions of the temporal and spatial pathways through a within-architecture ablation (Supplementary Text~5).

\begin{table}[h]
\caption{Per-lead MAE on the full 2021--2022 test set, reported as the mean across five random seeds and in normalized units (standardized over the training period). All DL models use the residual formulation, 90-day SM100 inputs, and a comparable parameter budget ($\sim$65\,M, except E3D-LSTM at 81\,M; see Supplementary Text~3); persistence and climatology (seed-independent) are shown for reference. The Mean row is the average over leads, with the across-seed standard deviation; the best value at each lead time is in bold. Verification is against ERA5. }\label{tab:perlead_mae}
\footnotesize\setlength{\tabcolsep}{3pt}
\begin{tabular}{@{}lcccccc@{}}
\toprule
Lead & SMCast & U-Net & CrossFormer & E3D-LSTM & Persistence & Climatology \\
\midrule
1  (d 1--5)   & \textbf{0.1035} & 0.1108 & 0.1130 & 0.1791 & 0.1081 & 0.3251 \\
2  (d 6--10)  & \textbf{0.1291} & 0.1359 & 0.1366 & 0.1930 & 0.1375 & 0.3237 \\
3  (d 11--15) & \textbf{0.1473} & 0.1545 & 0.1539 & 0.2072 & 0.1602 & 0.3240 \\
4  (d 16--20) & \textbf{0.1582} & 0.1673 & 0.1638 & 0.2194 & 0.1760 & 0.3241 \\
5  (d 21--25) & \textbf{0.1686} & 0.1774 & 0.1744 & 0.2325 & 0.1932 & 0.3241 \\
6  (d 26--30) & \textbf{0.1825} & 0.1897 & 0.1888 & 0.2495 & 0.2128 & 0.3240 \\
7  (d 31--35) & \textbf{0.1928} & 0.2013 & 0.1993 & 0.2635 & 0.2316 & 0.3236 \\
8  (d 36--40) & \textbf{0.1947} & 0.2047 & 0.2002 & 0.2695 & 0.2382 & 0.3238 \\
9  (d 41--45) & \textbf{0.2011} & 0.2111 & 0.2062 & 0.2823 & 0.2517 & 0.3244 \\
10 (d 46--50) & \textbf{0.2184} & 0.2286 & 0.2247 & 0.3058 & 0.2787 & 0.3251 \\
\midrule
Mean          & \textbf{0.1696 $\pm$ 0.0008} & 0.1781 $\pm$ 0.0065 & 0.1761 $\pm$ 0.0011 & 0.2402 $\pm$ 0.0266 & 0.1988 & 0.3242 \\
\botrule
\end{tabular}
\end{table}

Given that SMCast consistently outperforms the deep learning architectures considered here, the remainder of this section focuses on a detailed evaluation of SMCast, including its direct training variant, relative to established operational forecasting systems and standard statistical baselines.

\paragraph{S2S forecast skill against operational and statistical baselines}\label{subsec:skill}

\begin{figure}[h]
\centering
\includegraphics[width=0.9\textwidth]{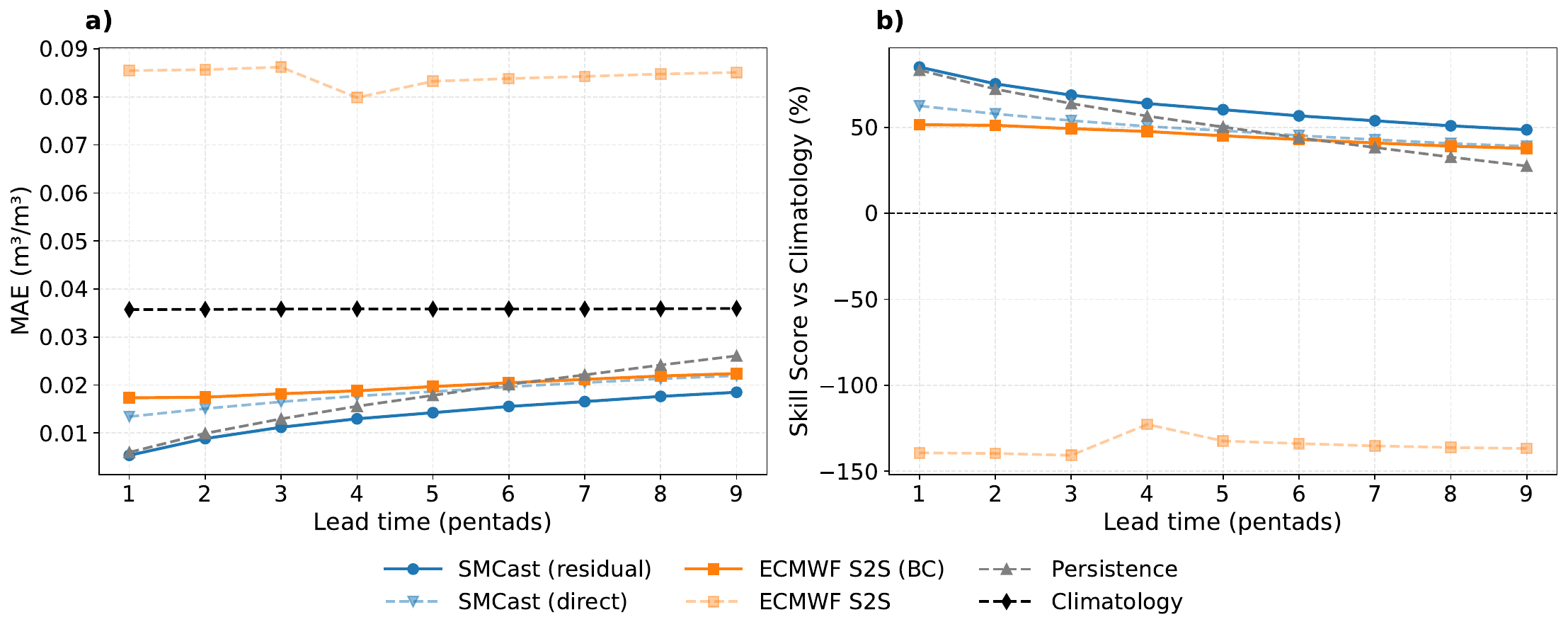}
\caption{Forecast error and skill against climatology across pentad lead times. a)~MAE (lower is better). b)~Skill score vs.\ climatology (\%; higher is better). Different colors lines and shapes correspond to the different models and baselines considered in this study. All scores are verified against ERA5 over the 2021--2022 test period and reported in physical units (m$^3$\,m$^{-3}$).}\label{fig:mae}
\end{figure}

We next assess the deterministic model performance in terms of standard metrics (see Section~\ref{subsec:metrics}). Note that the metrics are reported in physical units (m³/m³) consistent with the real-time S2S forecasts. Figure \ref{fig:mae} displays the mean absolute error (MAE) as a function of lead time ( the root mean squared error (RMSE) is additionally illustrated in Supplementary Fig.~S2). SMCast consistently outperforms all baselines, including persistence, climatology, and both the raw and BC ECMWF S2S forecasts, across all lead times. This advantage is particularly pronounced at longer lead times where persistent land-surface memory interacts with evolving atmospheric forcing \cite{Seneviratne2010}. 

The residual learning formulation is crucial for surpassing the persistence baseline, especially within the first two pentads, when soil moisture exhibits strong temporal persistence and long memory timescales associated with slow land-surface drying processes and evapotranspiration losses \citep{Ghannam2016}. When training the model via the \textit{direct} approach, we found that while the model still outperforms climatology and the raw ECMWF S2S model, SMCast struggles to beat the persistence during the first three pentads due to the strong soil moisture memory. We found that SMCast (residual) achieves the highest skill at every lead and overall (+62.6\%), with skill decreasing with lead time as all forecasts converge toward climatology at longer ranges (see right panels in Figure \ref{fig:mae}). Persistence (+52.0\%) and the bias-corrected ECMWF S2S (+45.0\%) display positive skill scores but consistently below SMCast, while the \textit{direct} variant (+48.9\%) is less skillful than the residual model, most notably at short leads where residual learning best exploits soil moisture memory. The raw ECMWF S2S is the only system that falls below the climatological reference across all leads, a direct consequence of its systematic wet bias (discussed below).

The lead-time-dependent performance shown in Figure \ref{fig:mae} is consistent with the aggregated results presented in Table \ref{tab:overall_performance}. The raw ECMWF S2S forecasts exhibit a systematic wet bias of approximately 0.05 m$^3$/m$^3$ relative to ERA5, resulting in higher MAE values that exceed even the climatological baseline. A pronounced temporal inhomogeneity is also visible around lead 4 (days 16--20), which is consistent with the resolution change from T639 to T319 in the ECMWF model at day 16 \citep{Lorenz2024}, as previously reported for shallow soil moisture \citep{Otero2025}. After QM bias correction, ECMWF S2S (BC) error is substantially reduced (averaging 0.0197 m$^3$/m$^3$), bringing its performance close to that of persistence. This highlights the effectiveness of statistical post-processing in mitigating systematic model biases, although performance remains below that of the top-performing data-driven approach.

\begin{table}[h]
\caption{Overall forecast performance averaged over the first nine pentad lead times (days 1--45). The comparison is restricted to leads 1--9 because the ECMWF S2S forecasts extend only to day 46, so the days 46--50 pentad is not available for S2S. Lower MAE/RMSE indicates better performance. All scores are verified against ERA5 over the 2021--2022 test period and reported in physical units (m$^3$\,m$^{-3}$). }\label{tab:overall_performance}
\begin{tabular}{@{}lcc@{}}
\toprule
Model & MAE (m$^3$/m$^3$) & RMSE (m$^3$/m$^3$) \\
\midrule
Climatology        & 0.0358 & 0.0498  \\
Persistence        & 0.0172 & 0.0266  \\
ECMWF S2S (raw)    & 0.0843 & 0.1234  \\
ECMWF S2S (BC)     & 0.0197 & 0.0304  \\
SMCast (direct)    & 0.0183 & 0.0263  \\
SMCast (residual)  & \textbf{0.0134} & \textbf{0.0209} \\
\botrule
\end{tabular}
\end{table}

\begin{figure}[H]
\centering
\includegraphics[width=\textwidth]{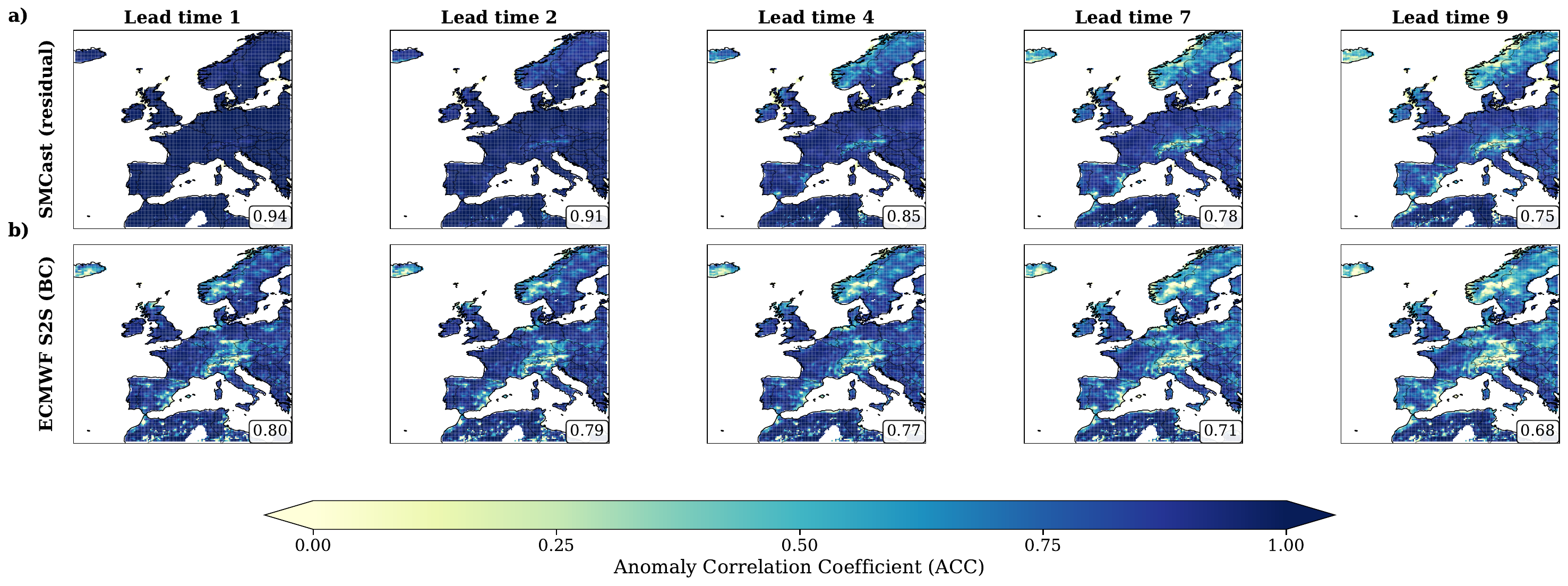}
\caption{Spatial anomaly correlation coefficient (ACC) for SMCast and ECMWF S2S (BC) at selected lead times. a)~SMCast (residual) at pentad leads 1, 2, 4, 7 and 9; b)~ECMWF S2S (BC) at the same leads. Higher values (darker blue) indicate better agreement with ERA5 anomalies. Anomalies are computed relative to the 1980--2015 training-period pentad-of-year climatology.}\label{fig:acc}
\end{figure}

We qualitatively analyze the spatial distribution of model performance through the anomaly correlation coefficient (ACC) and skill score (SS), i.e., as the relative improvement over climatology. For clarity, we present only the ECMWF S2S (BC) forecasts in the spatial comparison, as they consistently outperform the raw ECMWF S2S model across all metrics. As shown in Figure \ref{fig:acc}, SMCast achieves higher ACC values than the ECMWF S2S (BC) baseline across all lead times. As expected, forecast skill decreases with increasing lead time in both models, with the lowest ACC values observed over northern and central Europe. 

\begin{figure}[H]
\centering
\includegraphics[width=\textwidth]{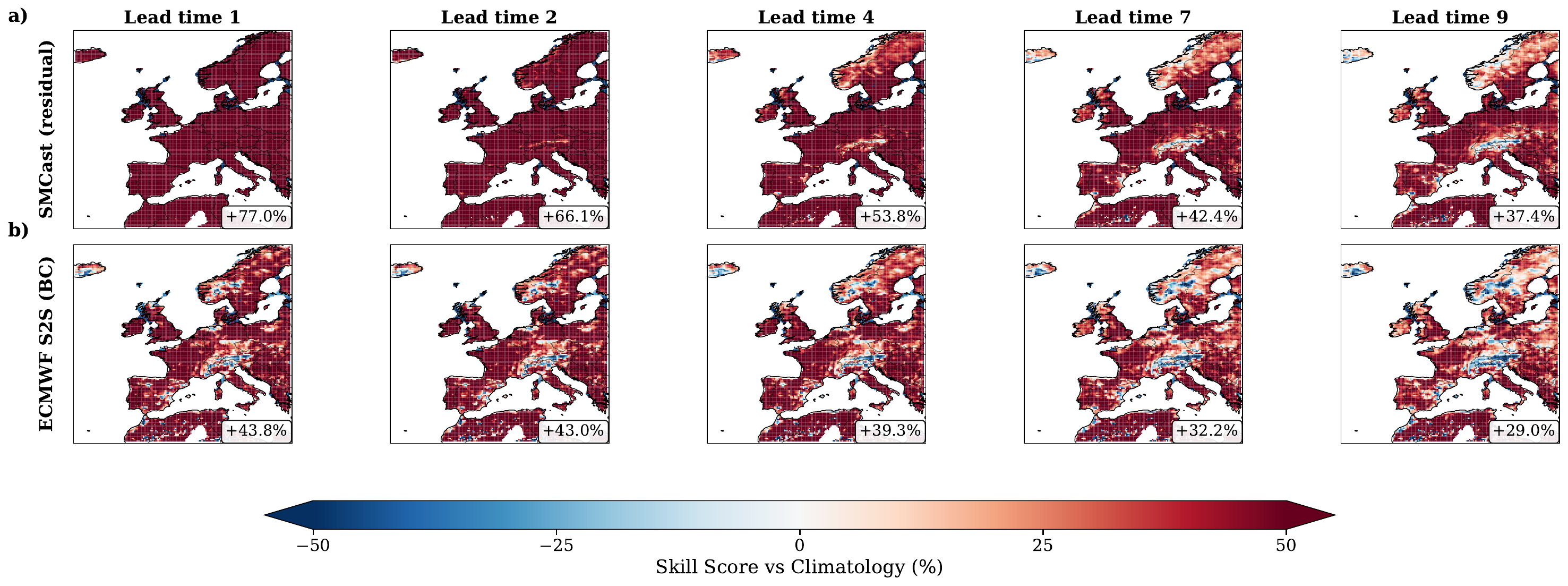}
\caption{Spatial skill score vs. climatology (\%) for SMCast and ECMWF S2S (BC) at selected pentad lead times (leads 1, 2, 4, 7, and 9; days 1--45). a)~SMCast (residual); b)~ECMWF S2S (BC), in the same lead order. Blue indicates regions where the model outperforms climatology; red indicates lower skill. Verification is against ERA5 over the 2021--2022 test period, with the reference climatology computed over the 1980--2015 training period.}\label{fig:ss_spatial}
\end{figure}

This spatial degradation pattern is further confirmed in Figure \ref{fig:ss_spatial}, which shows that the SS approaches zero in those regions at longer lead times (leads 7 and 9), indicating a loss of predictive advantage over climatology. In particular, central Europe exhibits a pronounced decline in skill for both models as forecast lead time increases, consistent with reduced soil moisture memory and increased influence of short-term meteorological variability in regions of complex orography \citep{Silvestri2025}.

\paragraph{Role of prediction formulation and target variable}\label{sec:target_choice}

Ablation experiments reveal that, in addition to the prediction formulation, the choice of target variable has a strong impact on forecast skill. Table~\ref{tab:ablation_perlead} compares the MAE across lead times for SM100 and SMA, both utilizing the residual configuration. Because SM100 and SMA are normalized independently, their absolute MAE values are not directly comparable; the relevant comparison is the skill score relative to each variable's own persistence baseline.

\begin{table}[!htbp]
\caption{Per-lead MAE (normalized units) for SM100 vs. SMA as target variable, both using the residual formulation with 90-day inputs. Persistence MAE is shown for reference. Mean MAE and SS relative to the Persistence baseline are reported on the bottom row. Verification is against ERA5 over the 2021--2022 test period; values are in normalized units (standardized with training-period statistics). Both configurations are single training runs (seed~42), so the values are not directly comparable with the five-seed means of Table~\ref{tab:perlead_mae}.}\label{tab:ablation_perlead}
\begin{tabular}{@{}lcccc@{}}
\toprule
 & \multicolumn{2}{c}{SM100 (residual)} & \multicolumn{2}{c}{SMA (residual)} \\
\cmidrule(lr){2-3} \cmidrule(lr){4-5}
Lead & MAE & Pers MAE & MAE & Pers MAE \\
\midrule
1  (d 1--5)   & 0.1035 & 0.1081 & 0.3112 & 0.3090 \\
2  (d 6--10)  & 0.1289 & 0.1375 & 0.4427 & 0.4411 \\
3  (d 11--15) & 0.1470 & 0.1602 & 0.5293 & 0.5280 \\
4  (d 16--20) & 0.1575 & 0.1760 & 0.5836 & 0.5826 \\
5  (d 21--25) & 0.1676 & 0.1932 & 0.6295 & 0.6284 \\
6  (d 26--30) & 0.1817 & 0.2128 & 0.6763 & 0.6753 \\
7  (d 31--35) & 0.1916 & 0.2316 & 0.7161 & 0.7153 \\
8  (d 36--40) & 0.1935 & 0.2382 & 0.7340 & 0.7332 \\
9  (d 41--45) & 0.1999 & 0.2517 & 0.7593 & 0.7586 \\
10 (d 46--50) & 0.2173 & 0.2787 & 0.8020 & 0.8014 \\
\midrule
Mean           & 0.1689 & 0.1988 & 0.6184 & 0.6173 \\
SS vs Pers     & +15.1\% &        & -0.2\% &        \\
\botrule
\end{tabular}
\end{table}

These two target representations exhibit markedly different behaviors. The SM100 residual model outperforms persistence at every lead, with a skill gain that increases from +4\% at lead 1 to +22\% at lead 10 (mean +15.1\%). In contrast, the SMA residual model fails to improve upon persistence under the same 90-day configuration (mean $-0.2\%$), effectively reproducing the persistence forecast across all leads. Sensitivity experiments further reveal contrasting responses to the input-window length. While SM100 benefits from extending the input window from 60 to 90 days, SMA is the only target representation whose performance deteriorates with a longer context. For SMA, skill decreases from +3.4\% with a 60-day window to $-0.2\%$ with a 90-day window, indicating that additional historical information does not translate into improved predictive performance. Even under its optimal configuration, however, SMA achieves only +3.4\% skill, falling substantially below the +14--15\% obtained with SM100. These differences can be attributed to the nature of the two representations. SM100 preserves both the climatological state and anomaly information, providing a richer signal from which the residual learning framework can extract predictable departures from persistence. In contrast, SMA is standardized and centered near zero, reducing both its dynamic range and the amount of large-scale structure available for residual correction. Consequently, the residual formulation becomes considerably less effective when applied to anomaly-based targets. Based on these results, SM100 is adopted as the primary target throughout this study, as it yields substantially stronger skill while still enabling percentile-based FD diagnostics through post-hoc transformation (Section~\ref{sec:target_choice}).

This finding is consistent with the results presented in \citet{Edris2026}, who reported that their DroughtFormer model skillfully predicted the timing and sign of soil moisture climate anomalies over Africa but struggled to capture their magnitude, a limitation that propagated into reduced skill in detecting FD intensification. Our results suggest that this difficulty may be partly inherent to the anomaly representation itself rather than a model-specific limitation, i.e., when anomalies are used directly as prediction targets, the reduced dynamic range and the removal of the climatological signal limit the model's ability to learn meaningful residuals.

\paragraph{Probabilistic forecasts}\label{sec:prob}

\begin{figure}[H]
\centering
\includegraphics[width=0.9\textwidth]{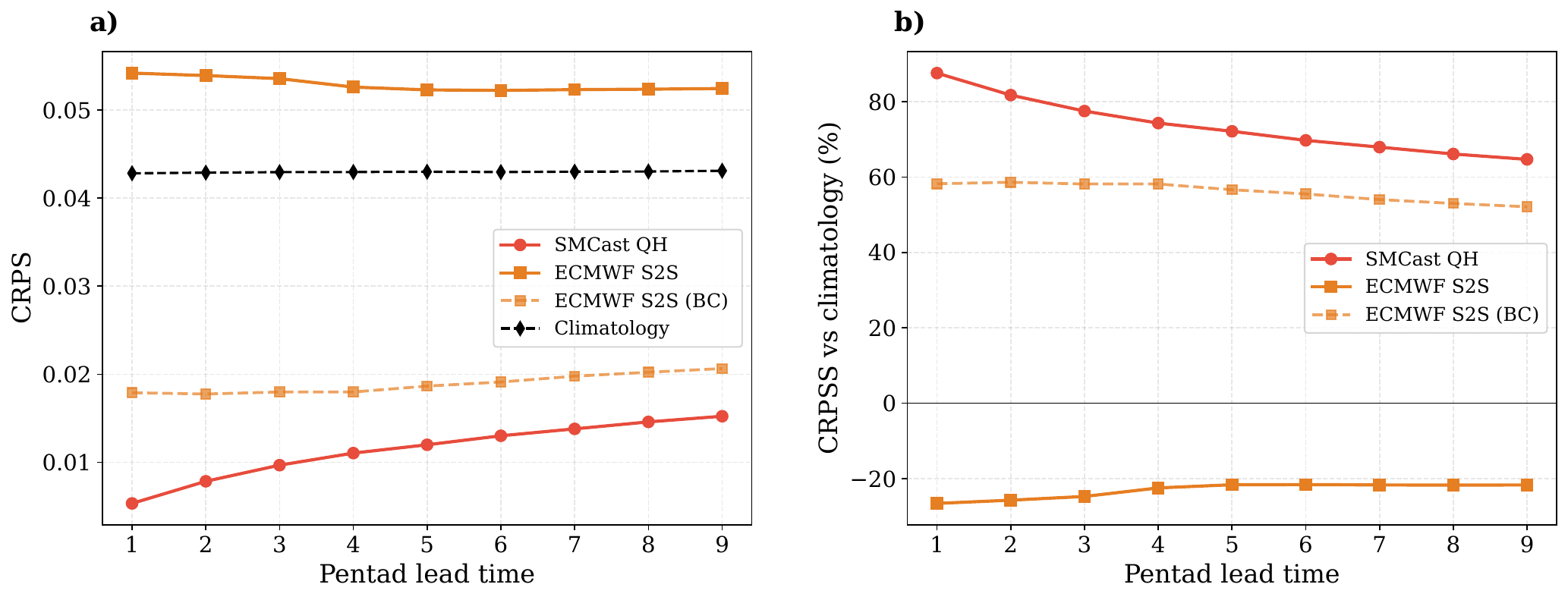}
\caption{Probabilistic forecast performance across pentad lead times. a)~CRPS per lead time in physical units (m$^3$/m$^3$) for SMCast QH, ECMWF S2S (raw), ECMWF S2S (BC), and climatology. b)~CRPSS vs climatology (\%). Verification is against ERA5 over the 2021--2022 test period, and the reference climatology is computed over the 1980--2015 training period.}\label{fig:crps}
\end{figure}

The probabilistic extension of SMCast by recalibrating the model via QH is evaluated using the Continuous Ranked Probability Score (CRPS). Figure~\ref{fig:crps} shows the CRPS and CRPSS relative to climatology across lead times. SMCast QH achieves substantially lower CRPS than both the raw and ECMWF S2S (BC) ensembles at all leads. Due to its systematic wet bias, the raw S2S ensemble performs worse than climatology (negative CRPSS). The bias-corrected S2S ensemble shows positive CRPSS but remains well below SMCast QH at all leads, confirming the superior performance of our approach.

Reliability diagrams (see Supplementary Fig.~S3) confirm that the SMCast QH quantiles are well calibrated, with observed frequencies closely following the diagonal across all quantile levels. The bias-corrected S2S ensemble, while improved over the raw ensemble, still exhibits miscalibration, particularly in the lower quantiles critical for drought monitoring.

\paragraph{Low-soil-moisture and flash drought evaluation}\label{sec:fd_results}

\begin{figure}[h]
\centering
\includegraphics[width=0.9\textwidth]{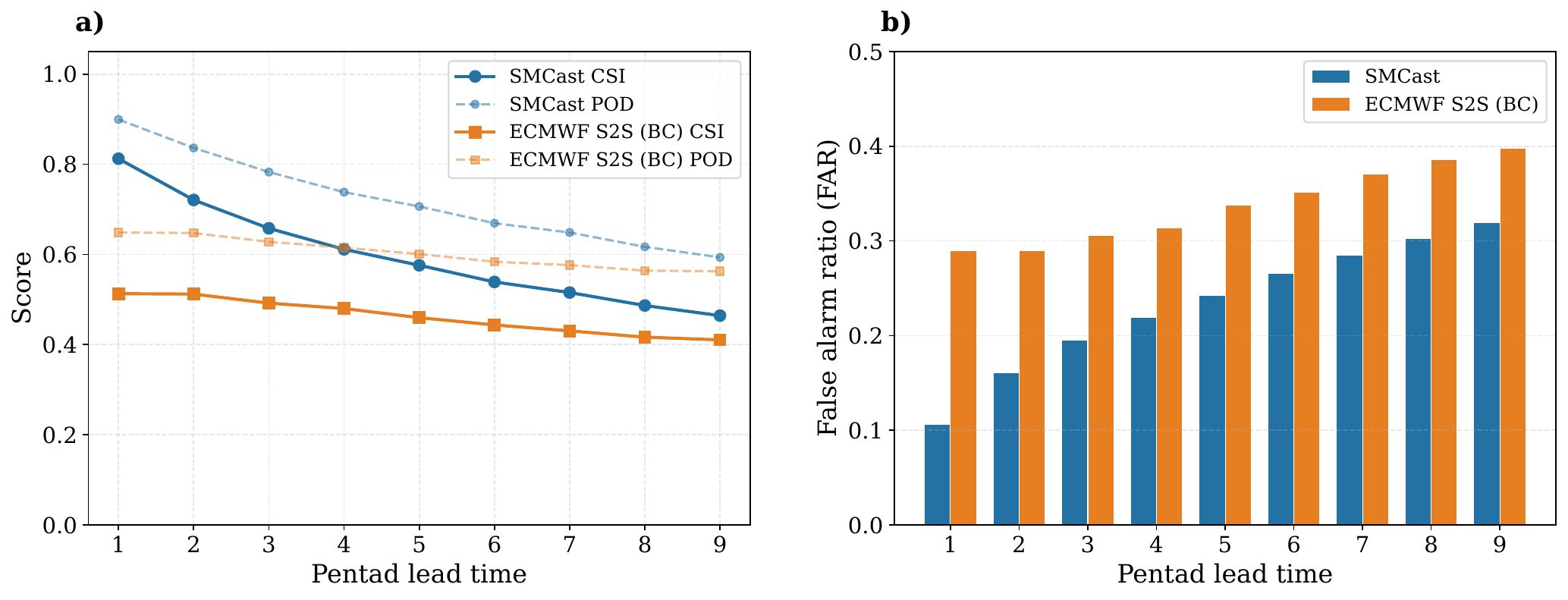}
\caption{Low-soil-moisture detection skill per lead time for the simple percentile threshold (SM percentile $<$ 20th), SMCast vs.\ ECMWF S2S (BC). a)~CSI and POD (higher is better): SMCast outperforms S2S~BC at all leads, with the largest advantage at short-to-medium leads. b)~FAR (lower is better): SMCast is consistently more precise than S2S~BC. The more demanding event-based definitions (Yuan onset, FDII) are reported in Table~\ref{tab:fd_onset}. Verification is against ERA5 percentiles derived from the 1980--2015 pentad-of-year climatology over the 2021--2022 test period.}\label{fig:fd_detection}
\end{figure}

We finally evaluate the ability of the model to support drought early warning using the post-hoc diagnostic framework, as described in Section~\ref{sec:fd_method}. Using a simple threshold approach, SMCast demonstrates strong skill in detecting low-soil-moisture conditions (SM percentile $<$ 20th) across all lead times (see Figure~\ref{fig:fd_detection}). At lead~1 (days 1–5), SMCast achieves a Critical Success Index (CSI) of 0.81 with a  Probability of Detection (POD) of 90\% and a False Alarm Ratio (FAR) of only 11\%, indicating that the model correctly identifies the vast majority of observed low-soil-moisture grid cells while producing minimal false alarms. As expected, skill degrades with lead time, reaching a CSI of 0.46, POD of 59\%, and FAR of 32\% at lead~9 (45~days). The ECMWF S2S (BC) forecasts start from substantially lower skill (CSI = 0.51 at lead~1) and converge toward SMCast only at the longest leads (CSI = 0.41 at lead~9). Nevertheless, SMCast is simultaneously more sensitive (higher POD) and more precise, maintaining a substantially lower false-alarm ratio than ECMWF S2S (BC) at every lead (Figure~\ref{fig:fd_detection}b), with FAR rising from 11\% to 32\% across the forecast window versus 29\% to 40\% for ECMWF S2S (BC). The advantage of SMCast for detecting extremes seems to be more pronounced at short-to-medium leads (pentads~1--6), where soil moisture memory provides additional predictability. 

This strong performance, however, is specific to the simple threshold definition. The onset criterion provides a large number of evaluation cases, yielding 318{,}618 FD-positive cases (grid cell $\times$ initialization) across pentad leads~5--9 of the 170 initializations common to ECMWF~S2S. A block-bootstrap analysis of these initializations gives 95\% intervals no wider than 0.04 in POD, indicating that the reported scores are not simply a consequence of having very few evaluation cases.
However, these cases were strongly correlated in space and time (see Discussion), so the limited number of independent FD episodes may still affect the generality of the reported skill.

The FD onset criteria of \citet{Yuan2019} yield modest binary detection skill (Table~\ref{tab:fd_onset}), with SMCast achieving CSI of 0.05 at lead~9 and a lower false alarm ratio than ECMWF S2S (BC) (FAR 0.81 vs 0.92), indicating higher precision when onset is predicted. While these scores are low in absolute terms, reflecting the tendency of the model to smooth rapid percentile transitions, they are consistent with the difficulty of capturing FD onset from bounded forecast windows reported across recent studies \citep{Edris2026, Zhao2025}. The FDII similarly shows low detection scores (CSI $<$ 0.05), though in this case the limitation is primarily structural, as the severity component integrates drought conditions over up to 18~pentads following onset, far exceeding the 10-pentad forecast horizon \citep{Otkin2018}. 

When evaluated as a continuous metric, SMCast achieves 28--36\% lower FDII MAE than ECMWF S2S (BC) across all leads where FDII events occur (Table~\ref{tab:fd_onset}), indicating that the predicted SM100 trajectories are closer to the observed intensification patterns. This is consistent with \citet{Edris2026}, who reported that their DroughtFormer struggled to accurately capture the rapid drop-off of moisture conditions necessary to predict FD even while producing skillful soil moisture forecasts.

\begin{figure}[h]
\centering
\includegraphics[width=0.9\textwidth]{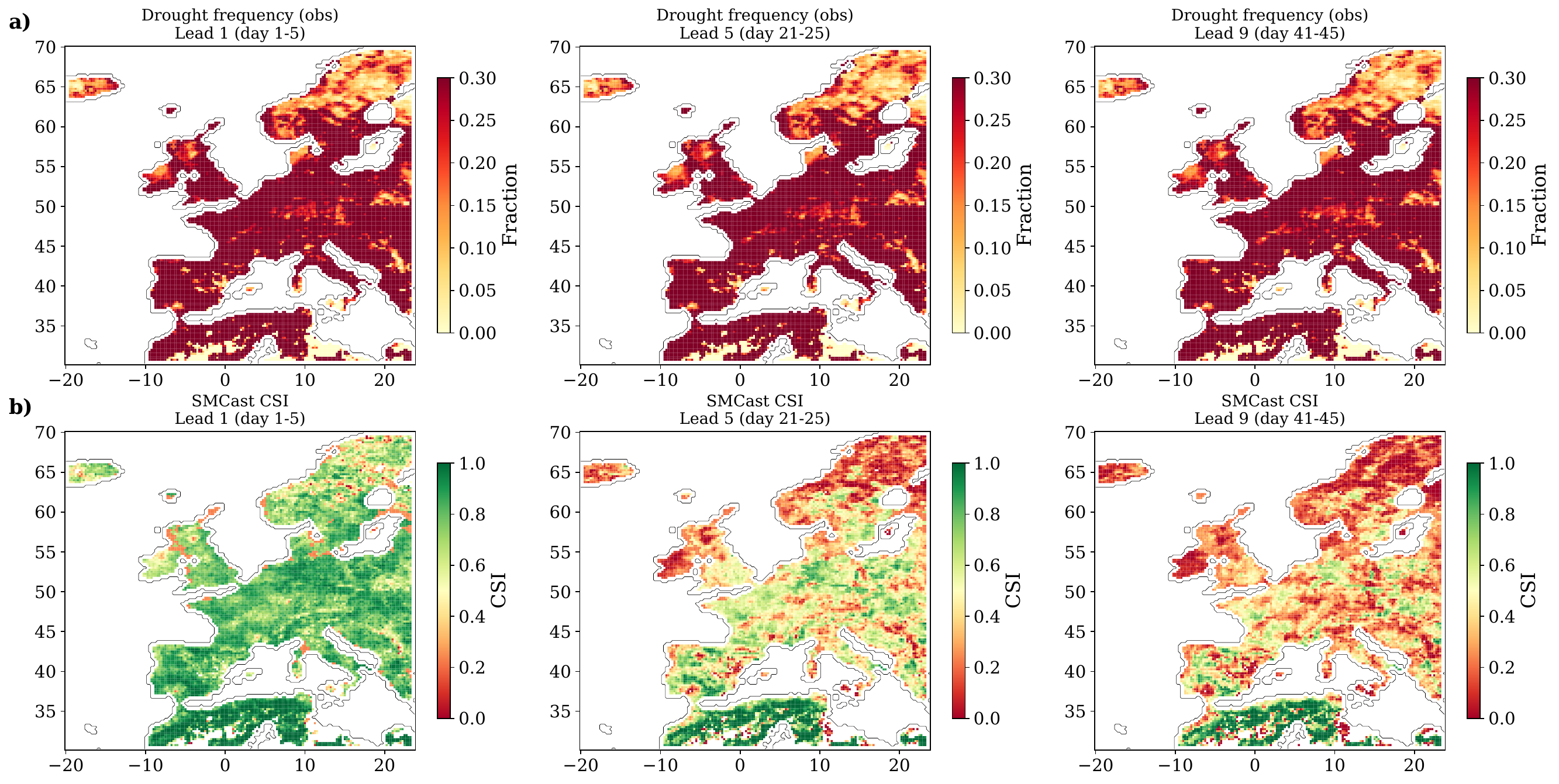}
\caption{Spatial low-soil-moisture detection over the 2021--2022 test period. a)~Top row: observed frequency of low-soil-moisture conditions (fraction of forecast dates with SM percentile $<$ 20th). b)~Bottom row: SMCast CSI at leads~1, 5, and~9. Verification is against ERA5 percentiles derived from the 1980--2015 pentad-of-year climatology, over the 2021--2022 test period.}\label{fig:fd_spatial}
\end{figure}

\begin{table}[h]
\caption{Event-based flash drought skill at selected lead times, for both SMCast (left) and ECMWF S2S (BC) (right). Both the Yuan onset criterion \cite{Yuan2019} and the FDII binary detection (FDII$>$0) yield near-zero CSI, reflecting the difficulty of resolving rapid intensification within a bounded forecast window. SMCast nonetheless shows higher precision (lower FAR) and lower continuous FDII error (FDII MAE) than ECMWF S2S (BC). Verification is against ERA5 percentiles derived from the 1980--2015 pentad-of-year climatology, over the 2021--2022 test period. }\label{tab:fd_onset}
\begin{tabular}{@{}lccccc@{}}
\toprule
 & \multicolumn{2}{c}{Yuan onset} & \multicolumn{2}{c}{FDII ($>$0)} & FDII MAE \\
\cmidrule(lr){2-3} \cmidrule(lr){4-5} \cmidrule(lr){6-6}
Lead & CSI & FAR & CSI & FAR & \\
\midrule
5  (d 21--25) & 0.04 / 0.02 & 0.84 / 0.95 & 0.02 / 0.02 & 0.88 / 0.95 & 0.24 / 0.35 \\
7  (d 31--35) & 0.05 / 0.03 & 0.80 / 0.93 & 0.04 / 0.04 & 0.85 / 0.92 & 0.42 / 0.66 \\
9  (d 41--45) & 0.05 / 0.04 & 0.81 / 0.92 & 0.04 / 0.05 & 0.85 / 0.90 & 0.33 / 0.46 \\
\botrule
\end{tabular}
\end{table}

Our results are generally consistent with recent studies showing that AI-based models can provide useful skill for FD prediction at S2S scales, although performance generally decreases rapidly with lead time due to the inherently fast onset and nonlinear evolution of flash drought events. For example, \citet{Zhao2025} reported F1 scores of approximately 0.83 at 1-day lead time for the IT-Drought Transformer, declining below 0.30 beyond three weeks, using a trajectory-based definition in which soil moisture drops below the 40th percentile and subsequently recovers above the 20th percentile. Similarly, \citet{Yang2026} reported hit rates of 71\% and FAR values of 20\% at 1-pentad lead time following the onset criteria proposed by \citet{Yuan2019}. 

The spatial distribution of detection skill (Figure~\ref{fig:fd_spatial}) reveals that CSI is highest in western and southern Europe, where the 2022 drought was most severe, and degrades over northern and eastern regions where dry conditions were less pronounced. However, we acknowledge that skill does not extend to flash drought onset detection, and under event-based onset criteria, CSI remains near zero for all systems evaluated, pointing to a fundamental limitation of fixed-horizon S2S forecasting for rapid intensification events (discussed in Section~\ref{sec:disc}).

\begin{figure}[h]
\centering
\includegraphics[width=0.5\textwidth]{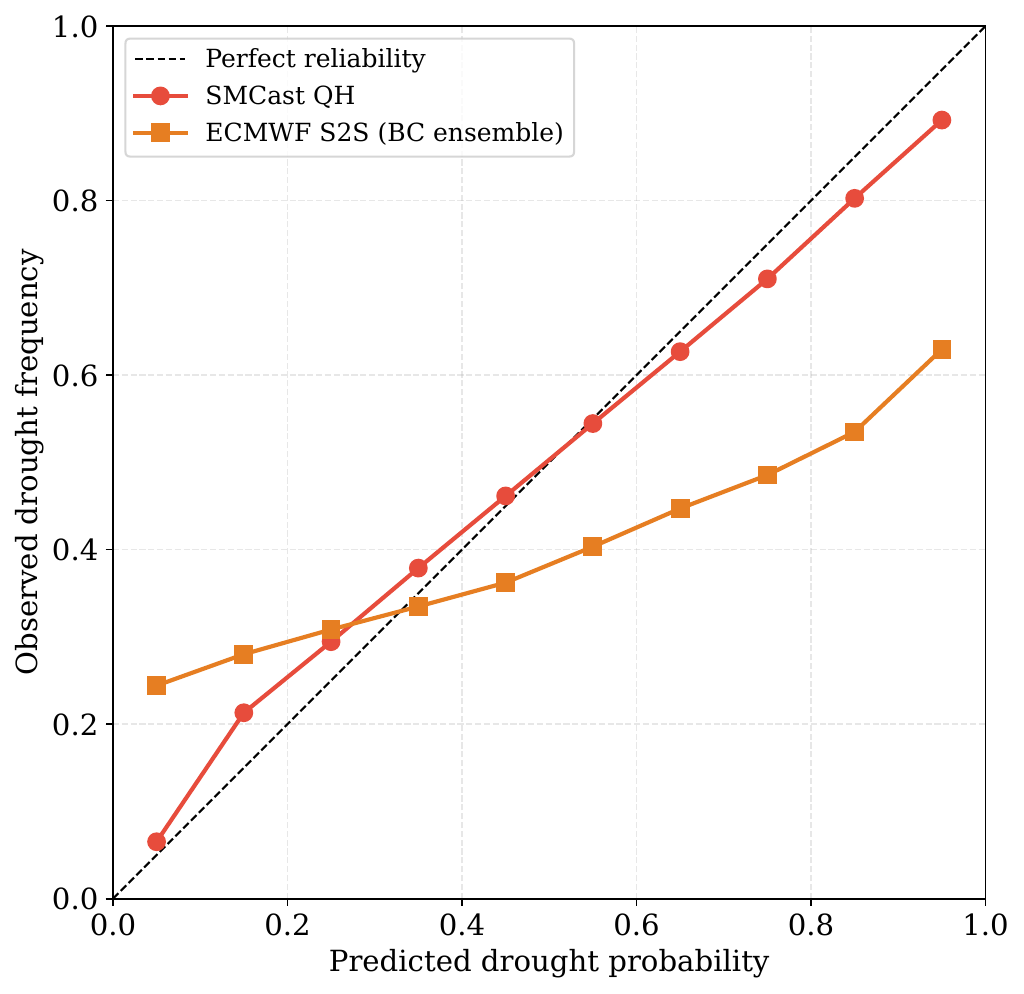}
\caption{Reliability diagram for the probabilistic detection of low-soil-moisture conditions (SM percentile $<$ 20th). SMCast QH (red) closely follows the diagonal, indicating well-calibrated low-soil-moisture probabilities. The ECMWF S2S (BC) ensemble (orange) is systematically overconfident. Verification is against ERA5 percentiles derived from the 1980--2015 pentad-of-year climatology, over the 2021--2022 test period.}\label{fig:fd_reliability}
\end{figure}

A well-calibrated forecast system should produce low-soil-moisture frequencies close to the climatological expectation. Over the 2021--2022 test period, we observed that ERA5 observations show 20.9\% of land grid cells below the 20th percentile, while SMCast predictions yield 19.6\% and ECMWF S2S (BC) yields 18.7\%. We conclude that this bias of SMCast is relatively small and consistent with the known tendency of regression-based deep learning models to underestimate the tails of the distribution \citep{Lesinger2025, Yang2026}. The empirical CDF of predicted percentiles confirms that all three systems show similar distributions, with SMCast closely tracking ERA5 across the full percentile range (see Supplementary Fig.~S4). This small but systematic conservative bias motivated the subsequent probabilistic evaluation. The SMCast QH provides calibrated probabilistic forecasts of low-soil-moisture conditions (Figure~\ref{fig:fd_reliability}). The reliability diagram shows that QH-predicted low-soil-moisture probabilities closely track the observed frequencies across all probability bins, with the calibration curve lying near the diagonal. In contrast, the ECMWF S2S (BC) ensemble exhibits more substantial miscalibration, and when the ensemble predicts a 90\% probability, low soil moisture is observed in only $\sim$63\% of cases. At an optimal probability threshold of $P > 0.4$, the SMCast QH low-soil-moisture forecasts achieve a mean CSI of 0.61 across leads 1--9. We emphasise that this diagnostic quantifies the probability of \emph{being} in a low-soil-moisture state, not the probability of satisfying the rapid-intensification and onset criteria of a flash drought; it therefore anticipates dry root-zone conditions rather than constituting a probabilistic forecast of flash-drought onset. This probabilistic capability is particularly relevant for operational early warning systems, where decision-makers require not only a binary classification but also a reliable estimate of forecast confidence \citep{Yang2026, Zhao2025}.

\section{Discussion}\label{sec:disc}

We have presented a data-driven framework for subseasonal soil moisture forecasting and FD early warning. Our results show that training formulation choices, in particular, the residual versus direct prediction strategies and the choice of target variable (absolute SM100 versus anomaly-based SMA) can dominate over architectural differences in determining forecast skill. A similar conclusion was reached by \citet{Garg2026}, who showed across ten state-of-the-art global AI weather models that training pipeline choices contribute as much to forecast skill as the underlying architecture.

In our study, we found that the forecast skill critically depends on the prediction formulation chosen for the soil moisture. In particular, the residual learning approach, which predicts deviations from the latest observed state rather than absolute soil moisture, is essential for outperforming persistence at short lead times, where strong temporal autocorrelation dominates predictability \citep{Ghannam2016, Seneviratne2010}. This formulation effectively removes the low-frequency persistence component and focuses the network's learning on meteorological forcing and land–atmosphere coupling. However, the residual formulation is not without trade-offs; the model exhibits a small conservative bias in percentile space, consistent with the tendency of regression-based models to underestimate distributional tails \citep{Yang2026, Lesinger2025}. This bias is especially relevant for drought applications, where tail behavior governs event detection. Incorporating a quantile head substantially improves calibration across the full distribution, indicating that probabilistic modelling has an added value to recover reliable extremes even when deterministic residual learning improves mean predictive skill. 

Furthermore, target representation further modulates the effectiveness of residual learning. SM100 consistently outperforms SMA, gaining +15.1\% over persistence, whereas SMA shows no skill over persistence ($-0.2\%$) under the same 90-day configuration. This contrast arises because anomalies are already centred and compressed, reducing the dynamic range available for residual learning, which limits the ability of the model to outperform the persistence baseline. This mechanism helps explain the challenges reported in \citet{Edris2026}, where their model reproduced anomaly timing but struggled to capture magnitude, reducing FD detection skill. These results point out that limitations previously attributed to model architecture may partly originate from the choice of target formulation. Predicting soil moisture in physical units and deriving drought indicators post hoc via percentile transforms provides a more robust formulation \citep{Zhao2025}. Future work should further explore how anomaly-based targets can be integrated without reducing predictive skill, for example, through hybrid formulations or multi-objective training that preserves both physical consistency and anomaly interpretability, as anomaly representations are central to S2S predictability and early warning systems \citep{He2025}.

SMCast compares favourably with recent data-driven approaches for S2S soil moisture and flash drought prediction. For instance, \citet{Zhao2025} introduced IT-Drought, a Transformer that forecasts daily soil moisture up to 46~days over the Continental United States (CONUS), reporting F1 scores that decline from 0.83 to below 0.30 beyond three weeks. Our pentad CSI of 0.81 at lead~1 is broadly consistent with their short-range skill, though different temporal resolutions, FD definitions, and domains preclude direct comparison. Unlike IT-Drought, SMCast employs factorized dual-pathway attention and residual learning, which may explain the maintained skill at longer leads. Similarly, \citet{Yang2026} employed an ensemble of six DL architectures with Bayesian model averaging over China, reporting hit rates of 71\% at 1-pentad lead but acknowledging conservative predictions under extreme conditions. Finally, \citet{Lesinger2025} demonstrated that combining recursive DL with dynamical model forecasts improves S2S soil moisture prediction, with the largest gains at weeks 3--4 where dynamical information becomes critical. Their hybrid DL-dynamical approach contrasts with our purely data-driven framework, suggesting that incorporating dynamical model information could further improve SMCast at extended leads. Across these studies, a common pattern emerges: all models exhibit a conservative bias for extreme events, and FD onset detection remains challenging regardless of architecture.

While our proposed framework improves S2S soil moisture prediction, several limitations remain. For instance, SMCast uses a direct multi-horizon decoder rather than an autoregressive approach. While this avoids error accumulation over the forecast window, it may limit the ability to capture rapid nonlinear transitions characteristic of flash drought onset. A recursive approach, as explored by \citet{Lesinger2025}, could improve longer-lead performance at the cost of accumulated errors. Moreover, the input-window analysis showed that skill saturates around a 90-day window: extending the inputs to 120 days provided no further gains at the medium-to-long lead times while increasing computational cost, suggesting that the temporal attention mechanism has difficulty extracting additional useful signal from very long sequences. We believe that more efficient long-range memory mechanisms may address this. Our evaluation relies on ERA5 reanalysis as both input and verification, meaning that systematic biases in ERA5 soil moisture propagate into both training and evaluation. Validation against in-situ observations or satellite products, such as the Global Land Evaporation Amsterdam Model (GLEAM) \citep{Miralles2025GLEAM4}, would strengthen confidence in real-world applicability. Finally, while the post-hoc FD evaluation demonstrates that skillful SM forecasts translate to useful low-soil-moisture detection, the low CSI ($\sim$0.04) under the event-based onset criteria highlights the fundamental difficulty of capturing rapid percentile transitions from smoothed forecast fields, a challenge shared across all recent studies \citep{Edris2026, Zhao2025}. This underscores that the main bottleneck in flash drought prediction is not overall soil moisture accuracy, but the ability to resolve the timing and magnitude of extreme transitions, which requires explicitly event-aware learning strategies. A further limitation is the relatively short test period, and a multi-year evaluation
covering more independent flash-drought episodes will be needed to assess the
generality of these results. Therefore, future work could explore asymmetric loss functions that penalize underestimation of extreme drying more heavily, or multiscale temporal attention that explicitly targets rapid intensification dynamics.

Despite these limitations, our work establishes a coherent framework for subseasonal soil moisture prediction and identifies clear directions for improving extreme-event representation and interpretability in data-driven early warning systems.

\section{Methods}\label{sec:methods}

\subsection{Data}\label{subsec:data}

We use the ERA5 reanalysis data \citep{Hersbach2020} from the European Centre for Medium-Range Weather Forecasts (ECMWF), which provides a physically consistent and temporally complete representation of the land--atmosphere system. All variables are processed at the original horizontal resolution of 0.25$^\circ$ and are subsequently aggregated at daily timescales over the period 1980--2022, covering Europe. Additionally, the soil moisture target variable is aggregated into pentads (5-day periods). This temporal resolution aligns with the timescales at which FD intensification is typically assessed \citep{Otkin2018, Tyagi2022} and provides a natural balance between capturing rapid soil moisture changes and filtering day-to-day variability that may add limited predictability at S2S scales.

The input features are selected to capture key processes governing land--atmosphere interactions and drought development. These include near-surface atmospheric conditions (2 m dew point temperature, daily maximum 2 m temperature, mean sea level pressure, and 10 m wind speed), surface fluxes (evaporation, potential evaporation, and latent and sensible heat fluxes), radiative forcing (surface net solar radiation), and precipitation (24-hour accumulated total precipitation). In addition, soil moisture conditions are represented through volumetric soil water content across four layers (0--7 cm, 7--28 cm, 28--100 cm, and 100--289 cm), enabling the model to capture both fast surface responses and deeper soil moisture memory. Table \ref{tabledata} summarizes the data used in this work.

Static fields, including geopotential at the surface and the land--sea mask, are incorporated to account for topographic and coastal contrasts that influence regional hydroclimate variability.

The primary target variable is root-zone soil moisture (hereafter referred to as SM100), computed as the thickness-weighted average of volumetric soil water content over the top 100~cm (integrating ERA5 layers 1--3: 0--7, 7--28, and 28--100~cm, with respective weights of 0.07/0.21/0.72). This formulation directly matches the target depth of the operational ECMWF S2S baseline. This variable is particularly suitable for drought prediction, as it reflects both short-term atmospheric forcing and longer-term hydrological memory \citep{Otkin2018, Tyagi2022}.

In addition, we evaluate an alternative key-FD target based on standardized soil moisture anomaly (SMA). The SMA is computed by normalizing soil moisture with respect to a smoothed pentad climatology derived from a reference period (1980--2015, corresponding to the training period). Specifically, for each grid point, a climatological mean and standard deviation are estimated for each pentad of the year using a rolling window to reduce sampling noise. The anomaly is then defined as the standardized deviation of daily soil moisture from this climatology, enabling a consistent comparison of anomalies across seasons and regions regardless of baseline local wetness.

\begin{table}[h]
\caption{Summary of input and target variables.}\label{tabledata}
\begin{tabular}{@{}llll@{}}
\toprule
\textbf{Type} & \textbf{Short name} & \textbf{Description} & \textbf{Resolution} \\
\midrule
\multirow{16}{*}{Inputs}
& d2m      & 2m dewpoint temperature & \multirow{16}{*}{0.25$^\circ$, daily} \\
& t2m\_max  & Daily maximum 2m temperature &  \\
& msl      & Mean sea level pressure &  \\
& ws10     & 10m wind speed &  \\
& e        & Evaporation &  \\
& pev      & Potential evaporation &  \\
& slhf     & Surface latent heat flux &  \\
& sshf     & Surface sensible heat flux &  \\
& ssr      & Surface net solar radiation &  \\
& tp       & Total precipitation &  \\
& swvl1    & Soil water layer 1 (0--7 cm) &  \\
& swvl2    & Soil water layer 2 (7--28 cm) &  \\
& swvl3    & Soil water layer 3 (28--100 cm) &  \\
& swvl4    & Soil water layer 4 (100--289 cm) &  \\
& z        & Geopotential &  \\
& lsm      & Land--sea mask &  \\
\midrule
\multirow{2}{*}{Target}
& sm100 & Root-zone soil moisture (integrated) & \multirow{2}{*}{0.25$^\circ$, pentad} \\
& SMA   & Standardized soil moisture anomaly &  \\
\botrule
\end{tabular}
\end{table}

\subsection{Problem formulation and model}\label{subsec:model}

Given a sequence of $T_{\text{in}}$ daily fields of surface variables over the European domain, the model predicts the target, e.g., SM100, at $T_{\text{out}}$ future pentad lead times (5-day averages), spanning approximately 50 days ahead (i.e., $T_{\text{out}} = 10$). In our default configuration, the model uses $T_{\text{in}} = 90$ days of daily inputs, providing sufficient temporal context to capture synoptic variability, land-surface memory, and seasonal trends relevant for subseasonal prediction (the sensitivity to this parameter is analyzed in Supplementary Text~1). Formally, given an input tensor $\mathbf{X} \in \mathbb{R}^{T_{\text{in}} \times C \times H \times W}$ with $C$ input channels and spatial dimensions $H \times W$, the model outputs $\hat{\mathbf{Y}} \in \mathbb{R}^{T_{\text{out}} \times 1 \times H \times W}$.

We propose SMCast, a Vision Transformer adapted for spatio-temporal forecasting through dual-pathway attention and lead-time conditioned decoding. The architecture consists of three stages: (i) patch embedding with spatial positional encoding, (ii) a dual-pathway encoder that jointly models temporal evolution and spatial interactions, and (iii) a lead-time conditioned decoder that generates forecasts for each target horizon (see Figure~\ref{fig:framework}). 

Each daily input frame is partitioned into non-overlapping patches and projected into a sequence of tokens following the standard Vision Transformer formulation \citep{Dosovitskiy2021}. Specifically, the European domain is cropped to $H \times W = 160 \times 176$ grid cells at $0.25^\circ$, and each daily frame with $C=16$ channels is divided into non-overlapping $16 \times 16$ patches, giving $N = (160/16) \times (176/16) = 10 \times 11 = 110$ spatial tokens per timestep. Each patch, a $16 \times 16 \times C$ block of grid cells and channels, is flattened and linearly projected to the token dimension $d = 1024$ by a single shared embedding layer implemented as a strided convolution, and a learnable spatial position embedding is added per token to encode location; temporal dependencies are handled within the attention mechanism rather than by the embedding. With $T_{\text{in}} = 90$ this yields $90 \times 110 = 9\,900$ tokens, on which the temporal pathway attends over $T_{\text{in}} = 90$ positions per spatial location and the spatial pathway over $N = 110$ tokens per timestep. 
%Each patch covers $4^\circ \times 4^\circ$, i.e.\ approximately $300 \times 450$~km at mid-latitudes.

The patch size of 16 was chosen to balance spatial resolution and computational cost.
The $160\times176$ domain is exactly divisible by 16, avoiding padding or partial
patches at the domain boundaries. In addition, reducing the patch size to 8 increases
the number of spatial tokens from 110 to 440 per timestep and substantially increases
the cost of spatial attention, which scales quadratically with the number of tokens.
The $16\times16$ configuration therefore provides a practical balance between spatial
resolution and computational efficiency, which is further supported by the patch-size
sensitivity experiment reported in Supplementary Text~4.

Tokenization removes attention-mediated interaction between grid cells within the same patch, since cells inside one $16 \times 16$ block are not separate tokens and cannot attend to one another. Note that their relationship is not discarded, but represented by the linear projections rather than by attention: the patch embedding is a full linear map over all $16 \times 16 \times C$ values in the block, so within-patch structure is encoded rather than averaged, and the decoder is a shared projection head that maps each token back to a full $16 \times 16$ output block per lead time, so within-patch detail is reconstructed rather than upsampled from a single value. The architecture therefore represents within-patch structure with a fixed, learned operator and between-patch structure with data-dependent attention. The residual formulation (Section~\ref{subsec:model}) further limits the cost of a coarse tokenization for this particular target: because the network predicts a \emph{departure} from the last observed field, fine-scale spatial structure that is stationary over the forecast window is inherited from $y_0$ at full $0.25^\circ$ resolution and never passes through the tokenization at all. The sensitivity of skill to patch size is examined in Supplementary Text~4.

\subsubsection{Spatiotemporal encoder}

The standard Vision Transformer encoder applies self-attention uniformly across all tokens, treating the flattened spatiotemporal sequence as unstructured. However, due to the strong temporal and spatial dependencies present in geophysical fields, driven by physical processes operating across multiple scales, we replace the standard ViT blocks with a stack of $L$ factorized spatio-temporal attention blocks that explicitly separate temporal and spatial reasoning, following the divided space--time attention design of \citet{Bertasius2021}. Unlike joint space--time attention, each block applies self-attention along the temporal and spatial axes separately and merges them through a single shared feed-forward layer, substantially reducing computational cost while retaining the ability to model dependencies along both axes.

Each of these blocks contains two parallel self-attention pathways. The \textit{temporal pathway} operates independently at each spatial patch: for patch $n$, the representations across all input timesteps are gathered into a temporal sequence $\mathbf{Z}^{\text{time}}_n = [\mathbf{z}_{1,n}, \ldots, \mathbf{z}_{T,n}] \in \mathbb{R}^{T_{\text{in}} \times d}$, and self-attention is applied along this axis. This allows the model to capture land-surface memory, persistence effects, and lead-time-dependent dynamics at each location. Temporal position is encoded through Rotary Position Embeddings (RoPE; \cite{Su2024}), which are applied per attention head and encode relative temporal distances without fixed positional embeddings. Encoding relative rather than absolute temporal distance is well suited to the problem, since soil-moisture memory depends on the time interval between an event and the forecast rather than its absolute position within the input window. It also makes the model less dependent on a fixed input-window length by avoiding a learned position-embedding table.

The spatial pathway operates at each timestep independently, applying self-attention across all spatial patches at a given time. This enables the model to represent large-scale hydroclimatic dependencies and regional co-variability patterns, such as those arising from atmospheric forcing or spatially coherent soil moisture anomalies.

The outputs of the two pathways are combined through a learnable fusion gate and residual connections, allowing the model to adaptively weight temporal and spatial information depending on the input context. We find that this configuration consistently improves skill over an \textit{un-gated} sum of the two pathways (see Supplementary Text~2 ), with the improvement most pronounced for skill against persistence at medium-to-long lead times. This factorized design reduces computational cost compared to full spatio-temporal attention (which scales quadratically with $T_{\text{in}} \times H \times W$) while preserving the ability to model interactions across both axes.

\subsubsection{Lead-time conditioned decoder}

The decoder produces forecasts for all ten pentad lead times in a single forward pass, rather than predicting one step at a time autoregressively. This direct multi-horizon approach avoids the error accumulation that is a well-known limitation of autoregressive methods in extended-range forecasting \cite{Bonavita2024}.

After encoding, the model extracts the representation from the last input timestep. This final-timestep latent state summarizes the full temporal context of the input sequence, which includes persistence, recent trends, and seasonal signals, through the preceding attention layers. For each target lead time, a distinct learnable embedding is added to this representation, conditioning the shared decoder on the prediction horizon. The conditioned representation is then processed through a shared multi-layer projection head that maps patch-level embeddings back to the spatial grid.

Sharing the decoder across lead times acts as an implicit regularizer and enables transfer of learned dynamics across horizons, while the horizon-specific embeddings ensure that each pentad receives an appropriately differentiated prediction. This is important because the nature of predictability may change substantially with lead time: at short ranges (days 1--5), skill is dominated by persistence and recent land-surface conditions, whereas at extended ranges (days 46--50), the model must increasingly rely on
large-scale climate patterns captured by the spatial attention pathway. Although the inputs are provided at daily resolution, the decoder directly produces pentad-aggregated soil moisture fields without requiring intermediate daily predictions.

\subsubsection{Residual learning formulation}

A critical component of our approach is the \textit{residual learning} formulation, in which the model predicts the change in soil moisture from the last observed value rather than the absolute target. Let $y_0 \in \mathbb{R}^{1 \times H \times W}$ denote the most recently observed soil moisture field (at initialization time $t_0$). During training, the target is transformed to:

$$\Delta \mathbf{y}_\ell = \mathbf{y}_\ell - y_0 \quad \text{for each lead time } \ell$$

and the model is trained to predict $\Delta \hat{\mathbf{y}}_\ell$. At inference, absolute predictions are recovered by:

$$\hat{\mathbf{y}}_\ell = \Delta \hat{\mathbf{y}}_\ell + y_0$$

This formulation reduces the learning difficulty by centering targets near zero, naturally represents persistence as a baseline, and aligns with successful approaches in weather forecasting and post-processing. As we show in the Results, this formulation is critical for achieving skill above persistence, as models trained in direct mode (predicting absolute values) consistently failed to beat persistence in the shorter lead times.

\subsubsection{Probabilistic recalibration}\label{sec:qh}

To enable probabilistic forecasts, we introduce a second training stage in which a QH is fine-tuned on top of the pretrained deterministic SMCast. In this stage, the encoder and decoder weights are frozen, and only the quantile head is trained. This allows the model to learn predictive uncertainty without modifying the underlying spatio-temporal representation.

The model outputs $Q=11$ conditional quantiles of soil moisture anomalies at each grid cell and lead time, at the levels $\tau \in \{0.05, 0.10, 0.20, 0.30, 0.40, 0.50, 0.60, 0.70, 0.80, 0.90, 0.95\}$. The quantile grid provides greater resolution in the lower tail ($\tau=0.05, 0.10, 0.20$), which is particularly relevant for drought applications, while the $\tau=0.20$ node directly corresponds to the threshold used for drought identification. Using a moderate number of quantiles provides a practical balance between distributional resolution and reliable training of the probabilistic head, while avoiding the additional complexity of learning many closely spaced quantiles and the associated risk of poorly trained tail estimates \citep{Cannon2018}. Moreover, because the number of head outputs scales linearly with $Q$, a compact quantile grid keeps the probabilistic head sufficiently small to be fine-tuned on the frozen backbone.

The 11 quantiles provide a discrete, non-parametric representation of the predictive
cumulative distribution function (CDF), with its resolution determined by the chosen
quantile levels \citep{Gneiting2011}. They should therefore not be interpreted as
11 samples, nor compared directly with the 50 members of the ECMWF ensemble.
Training is performed using a Weighted Quantile Loss (WQL), which applies the pinball
loss independently to each quantile and aggregates the resulting losses across all
quantile levels. This encourages the model to learn the distributional shape and
probabilistic uncertainty of the forecasts while preserving the residual formulation
of the deterministic model, thereby maintaining consistency between the point and
probabilistic forecasts.

\subsubsection{Training and implementation details}

We train all models using the AdamW optimizer, with a weight decay of 0.05 applied to the backbone and 0.01 to the decoder heads. To encourage faster task-specific adaptation, the decoder heads are trained with a learning rate three times higher than that of the backbone. The learning rate schedule consists of a linear warm-up over the first 5\% of training steps, followed by cosine annealing to zero.

For the deterministic model, we minimize the mean absolute error (MAE) over valid grid points (i.e., non-missing land values), which provides a robust objective for soil moisture prediction. To regularize training, we employ stochastic depth (drop path), with rates increasing linearly from 0 to a maximum value between 0.05 and 0.2 across transformer layers.

The probabilistic quantile model is trained separately in a second stage using the frozen deterministic backbone and the Weighted Quantile Loss (WQL), as described in the previous subsection. The same residual formulation is used during quantile fine-tuning, ensuring consistency between deterministic and probabilistic training stages.

The dataset is split into training (1980--2015), validation (2016--2020), and test (2021--2022) periods to ensure robust out-of-sample evaluation. All input variables are standardized using statistics computed over the training period. All models were trained on a single NVIDIA RTX 5090 GPU, with SMCast requiring approximately 7 hours per run, highlighting the modest computational footprint of the approach.

\paragraph{Design choices}

Several design decisions were motivated by systematic ablation experiments. In particular, we evaluated configurations of 60, 90, and 120 days of daily fields as a window input length. The 90-day window provided the best overall skill, with the largest gains concentrated at the medium-to-long lead times (pentads~4 and 6--10) that are most relevant for subseasonal prediction. Given that soil moisture exhibits memory timescales ranging from weeks to several months \citep{Seneviratne2010, Ghannam2016}, a 90-day window appears to capture the relevant land-surface memory and seasonal context, whereas a 60-day window may truncate part of this slow-varying signal. On the other hand, extending the window to 120 days yielded no further improvement, being marginally better than the 90-day skill only at the shortest leads while increasing the computational cost, likely because distant inputs provide little predictive information at S2S timescales. We therefore adopt $T_{\text{in}} = 90$ days as default; detailed results are reported in Supplementary Text~1 (Supplementary Table~S1).

We also found that the choice of target variable significantly influences model performance. We compare the performances when using both targets, SM100 against SMA, under the residual learning formulation. The model trained with SM100 yielded substantially stronger skill relative to persistence, whereas SMA, being already centered with reduced dynamic range, limited the effectiveness of the residual approach. Thus, we decided to use SM100 as the primary target throughout this study, with flash drought indicators derived post-hoc through percentile transformation (Section~\ref{sec:fd_method}). A quantitative comparison is presented in Section~\ref{sec:target_choice}.

\subsection{Model baselines and evaluation}\label{subsec:baselines}

We evaluate our model against two complementary sets of baselines. For the capacity-controlled comparison on the full test set, we benchmark SMCast against three deep-learning architectures with contrasting inductive biases. For a further comparison against operational forecasting and statistical baselines, we use four physical and statistical baselines, namely persistence, climatology, the operational ECMWF S2S forecasting system, and a bias-corrected version, as described below.

\paragraph{Deep learning baselines}
We additionally compare SMCast against three widely used deep learning architectures for spatio-temporal forecasting: a fully convolutional encoder--decoder (U-Net; \citep{Ronneberger2015}), a recurrent eidetic 3D-LSTM (E3D-LSTM; \citep{wang2018eidetic}), and a CrossFormer \citep{wang2022crossformer}, which is also the architecture underlying DroughtFormer \citep{Edris2026, schreck2025credt}. All baselines were adapted to our framework design and trained under the same residual learning formulation as SMCast, with identical input variables, 90-day input window and the same primary target variable (SM100). Crucially, the baselines were configured to a comparable parameter budget ($\sim$65\,M, except the recurrent E3D-LSTM at 81\,M; Supplementary Text~3), so that the comparison is \emph{capacity-controlled}: it isolates the contribution of distinct architectural priors, such as spatial convolutions, sequential recurrence, and cross-scale attention, against the factorized dual-pathway temporal/spatial attention used in SMCast, rather than reflecting differences in model size. Architecture-specific hyperparameters (depth, hidden dimensions, attention heads) and parameter counts are reported in Supplementary Text~3.

\paragraph{Persistence}
The persistence baseline assumes that the most recent observation remains constant over the forecast horizon. For all lead times, the prediction is equal to the last observed soil moisture value. This baseline is widely used in subseasonal-to-seasonal forecasting applications \citep{Nathaniel2024, Mouatadid2023, Rasp2024}, particularly due to the strong short-term temporal autocorrelation of soil moisture.

\paragraph{Climatology}
The climatology represents the long-term seasonal mean computed from the training dataset (1980--2015). For each grid point and calendar pentad, predictions correspond to the average soil moisture for that time of year. In this study, climatology is computed by grouping the training data by pentads derived from the day of year (1--73). Leap days (February 29) are assigned to pentad 11 to preserve seasonal consistency. For each pentad and grid cell, we compute both the mean and standard deviation of soil moisture over the training period. These statistics define the reference seasonal cycle and are used for evaluating anomalies and model skill.

\paragraph{ECMWF S2S forecast model}
We further compare against the operational Subseasonal-to-Seasonal (S2S) forecasting system from the European Centre for Medium-Range Weather Forecasts (ECMWF), which provides physically based ensemble forecasts initialized from reanalysis conditions and simulates the coupled atmosphere--land--ocean system \citep{Vitart2018}.

We use the ECMWF S2S model, which produces both reforecasts and real-time forecasts, issued twice weekly (Mondays and Thursdays). The system provides predictions with lead times of up to 46 days at a spatial resolution of approximately $1.5^\circ$, with output available at daily or 6-hourly frequency depending on the variable. In this study, only real-time forecasts are considered for the evaluation. Specifically, we use forecasts from 2021 and 2022 comprising 51 ensemble members (50 perturbed members and one control member), resulting in a total of 208 forecast initializations. To ensure consistency with our deterministic framework, we discard the control member and use the ensemble mean of the perturbed members. Daily averages of soil moisture in the top 100 cm are retrieved and subsequently aggregated to pentad scale. As the native forecasts are provided at coarse spatial resolution, they are bilinearly interpolated onto the common ERA5 grid ($0.25^\circ$), rather than the $0.25^\circ$ fields being aggregated to $1.5^\circ$. We decided to adopt the common $0.25^\circ$ grid because the verification target and all drought diagnostics, including the 20th-percentile threshold and FD metrics, are defined from the $0.25^\circ$ ERA5 climatology. Aggregating to the native $1.5^\circ$ ECMWF grid would smooth spatial variability and alter these diagnostics, while also making the ECMWF
comparison inconsistent with SMCast and the other $0.25^\circ$ baselines. The ECMWF forecasts are therefore regridded to $0.25^\circ$ before quantile mapping, which is performed for each grid cell and lead time against ERA5. This does not add spatial information to the ECMWF forecasts, but provides a common evaluation grid and accounts for systematic biases. This baseline represents a state-of-the-art physically constrained forecasting system and provides a strong benchmark for evaluating data-driven approaches.

Additionally, we evaluate a bias-corrected version of the ECMWF S2S forecasts using quantile mapping (QM), a widely used bias-correction technique that aligns the distribution of forecasts with that of observational reference data by matching their empirical quantiles \citep{Monhart2018}. Unlike mean bias-correction methods, QM adjusts the entire marginal distribution, thereby correcting both the frequency and intensity of the target variable \citep{Maraun2016}. QM has been widely used as a statistical post-processing baseline in subseasonal forecasting studies \citep{Mouatadid2023, Springenberg2026}.

We apply empirical QM independently at each grid point and lead time using ERA5 soil moisture as reference data. The mapping is estimated from discrete empirical quantiles (20 evenly spaced quantile levels) of the forecast and reference distributions and applied via interpolation between forecast and reference quantiles. The mapping is estimated on the ensemble-mean S2S--ERA5 pairs and applied both to the ensemble mean, giving the deterministic bias-corrected forecast, and unchanged to each of the 50 perturbed members, giving the bias-corrected ensemble on which all probabilistic S2S scores are based. Given the limited number of available real-time forecasts (208 samples), this quantile discretization provides a practical compromise between distributional resolution and sample robustness. To avoid temporal leakage, we adopt a two-fold cross-validation strategy in which mappings calibrated on one year are applied to the other year and vice versa. For each forecast valid time, the corresponding ERA5 target is obtained from the nearest pentad observation. This bias-corrected configuration serves as a statistical post-processing baseline for comparison with the raw ECMWF S2S forecasts.

\subsection{Low-soil-moisture and flash drought identification}\label{sec:fd_method}

To evaluate the capacity of SMCast forecasts for FD early warning, we assess both deterministic and probabilistic detection of low-soil-moisture conditions using a post-hoc diagnostic framework applied to predicted SM100 fields. Thus, for each grid cell and forecast valid time, predicted and observed SM100 values are converted to climatological percentiles using the pentad-of-year distribution from the training period (1980--2015). As a primary diagnostic, we evaluate the detection of low-soil-moisture conditions, defined as SM100 percentiles below the 20th percentile \citep{Lesinger2025}. In addition, we apply the FD identification criteria following \citep{Yuan2019}, which require SM100 percentiles to decline from above the 40th to below the 20th percentile at a rate exceeding 5~percentile points per pentad, sustained for at least four pentads, capturing both the rapid intensification and drought persistence components \citep{Yuan2019}. We distinguish these two diagnostics throughout: \emph{low-soil-moisture detection} refers to the 20th-percentile state threshold, whereas \emph{flash-drought detection} refers to the event-based rapid-intensification criteria, which are the more demanding target relevant to flash-drought early warning.

The Flash Drought Intensity Index (FDII; \citep{Otkin2018}) was also considered, though we note that this index requires look-back and look-forward windows that extend beyond the 10-pentad forecast horizon, limiting its applicability in our fixed-window S2S evaluation setting. Moreover, we initially explored a multitask formulation jointly predicting SM100 and FDII components, but this did not improve FD representation, consistent with the challenges reported by \citet{Edris2026}. Therefore, the results presented here are based on the post-hoc evaluation framework unless otherwise stated.

To complement the deterministic evaluation, we derive probabilistic low-soil-moisture forecasts from the SMCast QH version by computing the fraction of predicted quantiles falling below the 20th-percentile threshold, providing a spatially explicit probability of a low-soil-moisture state, $P(\mathrm{SM}<20\mathrm{th})$, at each lead time. This is compared against the ensemble-derived probability from the ECMWF S2S (BC) forecasts (50 members).

\subsection{Performance metrics}\label{subsec:metrics}

We evaluate model performance using standard deterministic verification metrics, including mean absolute error (MAE), root mean squared error (RMSE) and anomaly correlation coefficient (ACC). All verification uses ERA5 as the reference. As stated in section \ref{sec:methods} root-zone soil moisture (SM100) is aggregated to pentad means on the native $0.25^\circ$ grid, with ocean and missing
cells excluded. Scores are computed over the held-out 2021--2022 test period for each
lead time. All climatologies and percentile transforms are derived exclusively from the
1980--2015 training period. ECMWF~S2S metrics are reported in physical units, while
the deep-learning and ablation results use normalized units. 

To evaluate the probabilistic forecasts produced by the SMCast QH, we compute the
Continuous Ranked Probability Score (CRPS), which measures the distance between the
predicted CDF and the observed outcome \citep{Rasp2024}. Given that our model produces a
discrete set of quantiles, we approximate the predictive distribution using linear
interpolation between quantile levels \citep{Zamo2018}. The same score computed from the
50 S2S members is likewise a finite-sample estimator whose value depends on ensemble size
\citep{Hersbach2000,Ferro2008}. CRPS is then computed at each grid point and lead time and aggregated over space and time. Lower CRPS values indicate better calibrated and sharper predictive distributions. This metric complements deterministic scores such as MAE by explicitly evaluating the quality of uncertainty estimates. In addition, we assess calibration using reliability diagrams derived from the predicted quantiles, as well as prediction interval coverage for selected quantile ranges (e.g., 10--90\%). We compute the skill score (SS) relative to a reference forecast (i.e., climatology) as:

\begin{equation}
\mathrm{SS} = 1 - \frac{\mathrm{S}_{\mathrm{model}}}{\mathrm{S}_{\mathrm{ref}}},
\end{equation}
where $\mathrm{S}_{\mathrm{ref}}$ denotes the score (error metric) of the reference forecast. Positive values indicate improved performance over the reference, while negative values indicate degradation.

We evaluate categorical FD event-based skill using the Probability of Detection (POD), False Alarm Ratio (FAR), and Critical Success Index (CSI). These metrics assess the ability of the models to correctly identify FD occurrence while accounting for missed events and false alarms. Metrics are computed independently at each lead time using binary FD masks derived from the identification criteria described above and spatially aggregated over the evaluation domain.

\backmatter

\bmhead{Supplementary information}

Supplementary material includes the sensitivity analysis to the input window length (Supplementary Text~1, Supplementary Table~S1), the architectural ablation of RoPE and the fusion gate (Supplementary Text~2, Supplementary Table~S2, Supplementary Fig.~S1), the deep-learning baseline architectures and parameter counts (Supplementary Text~3, Supplementary Table~S3), the sensitivity to the spatial patch size (Supplementary Text~4, Supplementary Table~S4) and the ablation of the temporal and spatial attention pathways (Supplementary Text~5, Supplementary Table~S5). Additional tables and figures include the RMSE per lead time (Supplementary Table~S6 and Fig.~S2), the reliability diagram of the probabilistic forecasts (Fig.~S3) and the empirical CDF of predicted percentiles (Fig.~S4).

\bmhead{List of Abbreviations}

\begin{description}
\item[ACC] anomaly correlation coefficient
\item[BC] bias-corrected
\item[CDF] cumulative distribution function
\item[CRPS] continuous ranked probability score
\item[CRPSS] continuous ranked probability skill score
\item[CSI] critical success index
\item[DL] deep learning
\item[E3D-LSTM] eidetic 3D long short-term memory
\item[ECMWF] European Centre for Medium-Range Weather Forecasts
\item[ERA5] fifth-generation ECMWF atmospheric reanalysis
\item[F1] F1 score, the harmonic mean of precision and recall
\item[FAR] false alarm ratio
\item[FD] flash drought
\item[FDII] flash drought intensity index
\item[FG] fusion gate
\item[GLEAM] Global Land Evaporation Amsterdam Model
\item[MAE] mean absolute error
\item[ML] machine learning
\item[NWP] numerical weather prediction
\item[POD] probability of detection
\item[QH] quantile head
\item[QM] quantile mapping
\item[RMSE] root mean squared error
\item[RoPE] rotary position embeddings
\item[S2S] subseasonal-to-seasonal
\item[SM] soil moisture
\item[SM100] root-zone soil moisture, top 100~cm
\item[SMA] standardized soil moisture anomaly
\item[SMCast] soil moisture forecasting model proposed in this study
\item[SS] skill score
\item[ViT] Vision Transformer
\item[WQL] weighted quantile loss
\end{description}

\bmhead{Acknowledgements}
The authors would like to thank the two anonymous reviewers and the editor for their thoughtful and
detailed comments, which helped to improve the final version of this paper. This research has been supported by the European Union’s Horizon Europe research and innovation program (EU Horizon Europe) project MedEWSa under grant agreement no.101121192 and the ARTEMIS project under grant Agreement n.101188009. M.-Á. F.-T. thanks the Universidad Carlos III de Madrid, Spain, for the support provided through the Funding for Research Activities Program (project 2025/00803/001) and the Regional Government of Madrid through project TEC-2024/COM-322, and also acknowledges the computer resources provided by Artemisa, funded by the ERDF and Comunitat Valenciana, as well as the technical support provided by the Instituto de Física Corpuscular, IFIC (CSIC-UV).

\section*{Declarations}

\begin{itemize}
\item \textbf{Author contributions.} N.O. conceived the study, designed the methodology, developed the model code, performed the analysis, and wrote the original draft. A.\"O. contributed to model development and software implementation. M.-\'A.F.-T. discussed the methodology and the interpretation of results. J.M. acquired funding and administered the project. All authors reviewed and edited the manuscript and approved the final version.

\item \textbf{Competing interests.} The authors declare no competing interests.
\item \textbf{Data availability.} ERA5 reanalysis data are available through the Copernicus Climate Data Store (\url{https://cds.climate.copernicus.eu}). ECMWF S2S forecast data are available through the S2S Project database (\url{https://apps.ecmwf.int/datasets/data/s2s/}).

\item \textbf{Code availability.} The source code for SMCast, training scripts, and trained model checkpoints are available via Zenodo at https://doi.org/10.5281/zenodo.23105488 (version 1.0.0).

\end{itemize}

\bibliographystyle{unsrtnat}
\bibliography{references}

\end{document}